\documentclass{article}

\usepackage[final]{style}

\usepackage{amsmath,amsfonts,bbm}
\makeatletter
\g@addto@macro\normalsize{%
  \setlength\abovedisplayskip{4pt}
  \setlength\belowdisplayskip{4pt}
  \setlength\abovedisplayshortskip{4pt}
  \setlength\belowdisplayshortskip{4pt}
}
\usepackage{graphicx}
\usepackage{wrapfig}
\usepackage{float}
\usepackage[font={small},skip=1pt]{caption}
\usepackage[colorinlistoftodos]{todonotes}
\usepackage{mathtools}
\usepackage{xcolor}
\usepackage{natbib}
\usepackage{verbatim}
\usepackage{tikz}
\usepackage{enumitem}  
\usepackage{cancel}
\usepackage[ruled,vlined]{algorithm2e}  
\usepackage{algpseudocode}
\usepackage[user,titleref]{zref}
\usepackage{cancel}
\usepackage{hyperref}
\usepackage{gensymb}

\DeclareMathOperator*{\diag}{diag}

\newcommand{\bx}{\pmb{x}}
\newcommand{\bw}{\pmb{w}}

\newcommand{\br}{\pmb{r}}

\newcommand{\bphi}{\pmb{\phi}}

\newcommand{\bxi}{\pmb{\xi}}
\newcommand{\btau}{\pmb{\tau}}

\newcommand{\R}{\mathbb{R}}

\definecolor{darkgreen}{RGB}{40, 150, 40}
\setlist[itemize]{itemsep=0mm, topsep=2pt}
\setlist[enumerate]{itemsep=0mm, topsep=2pt}

\newcommand{\eq}[1]{\begin{align*}#1\end{align*}}

\title{Bayesian Optimization in Variational Latent Spaces with Dynamic Compression}

\author{
  Rika Antonova\thanks{Both of these authors contributed equally.\vspace{-5px}}  $\quad $ \\
  KTH Royal Institute of Technology, Sweden \\
  \texttt{antonova@kth.se} \\
  \And
  Akshara Rai\footnotemark[1] $\quad \quad \quad \quad \quad $ \\
  Facebook AI Research  $\quad \quad \quad \quad $ \\
  \texttt{akshararai@fb.com} $\quad \quad \quad \quad $\\
  \AND
  $\quad \quad \quad $ Tianyu Li\\
  $\quad \quad \quad $ Facebook AI Research \\
  \And
  Danica Kragic\\
  KTH Royal Institute of Technology, Sweden
}

\begin{document}
\maketitle

\vspace{-20px}
\begin{abstract}
Data-efficiency is crucial for autonomous robots to adapt to new tasks and environments. In this work we focus on robotics problems with a budget of only 10-20 trials. This is a very challenging setting even for data-efficient approaches like Bayesian optimization (BO), especially when optimizing higher-dimensional controllers. 
Simulated trajectories can be used to construct informed kernels for BO. However, previous work employed supervised ways of extracting low-dimensional features for these. 
We propose a model and architecture for a sequential variational autoencoder that embeds the space of simulated trajectories into a lower-dimensional space of latent paths in an unsupervised way. We further compress the search space for BO by reducing exploration in parts of the state space that are undesirable, without requiring explicit constraints on controller parameters.
We validate our approach with hardware experiments on a Daisy hexapod robot and an ABB Yumi manipulator. We also present simulation experiments with further comparisons to several baselines on Daisy and two manipulators. Our experiments indicate the proposed trajectory-based kernel with dynamic compression can offer ultra data-efficient optimization.
\end{abstract}



\vspace{-10px}
\section{Introduction}

Reinforcement learning (RL) is becoming popular in robotics, since
in some cases it can deal with real-world challenges, such as noise in control and measurements, non-convexity and discontinuities in objectives. However, most flexible RL methods require thousands to millions of data samples, which can make direct application to real-world robotics infeasible. 
For example, 10,000 30s trials/episodes on a real robot would require $\approx$100 hours of operation. Most full-scale platforms, especially in locomotion, cannot operate this long without maintenance. 
Nowadays, commercially available arms can operate for longer, however sophisticated anthropomorphic hands and advanced grippers are still highly prone to breakage after even a handful of trials~\cite{openai2018learning}. Hence the need for algorithms that can learn in very few trials, without causing significant wear-and tear to the hardware. 

In this work we focus on cases with a budget of only 10-20 trials. In such settings, using approaches like Bayesian optimization (BO) to adjust parameters of structured controllers can help improve data efficiency. However, success of BO on hardware has been demonstrated either with low-dimensional controllers or with simulation-based kernels that required hand-designed features.
We propose learning simulation-based kernels in unsupervised way with a sequential variational autoencoder (SVAE). Our approach embeds simulated trajectories $\bxi$ to a space of latent paths $\btau$, and jointly learns a probability distribution $p(\btau|\bx)$ that controllers with parameters $\bx$ induce over the space of latent paths. Our work is inspired by initial success of trajectory-based BO kernels~\cite{wilson2014using}, however that was demonstrated for BO in low dimensions (2-4D). Our results show that performance of a kernel based on raw trajectories deteriorates quickly for higher-dimensional problems. In contrast, a kernel based on latent paths can still offer gains even for 48-dimensional controllers. 

\begin{figure}[t!]
    \centering
    \vspace{-20px}
    \includegraphics[width=0.7\textwidth]{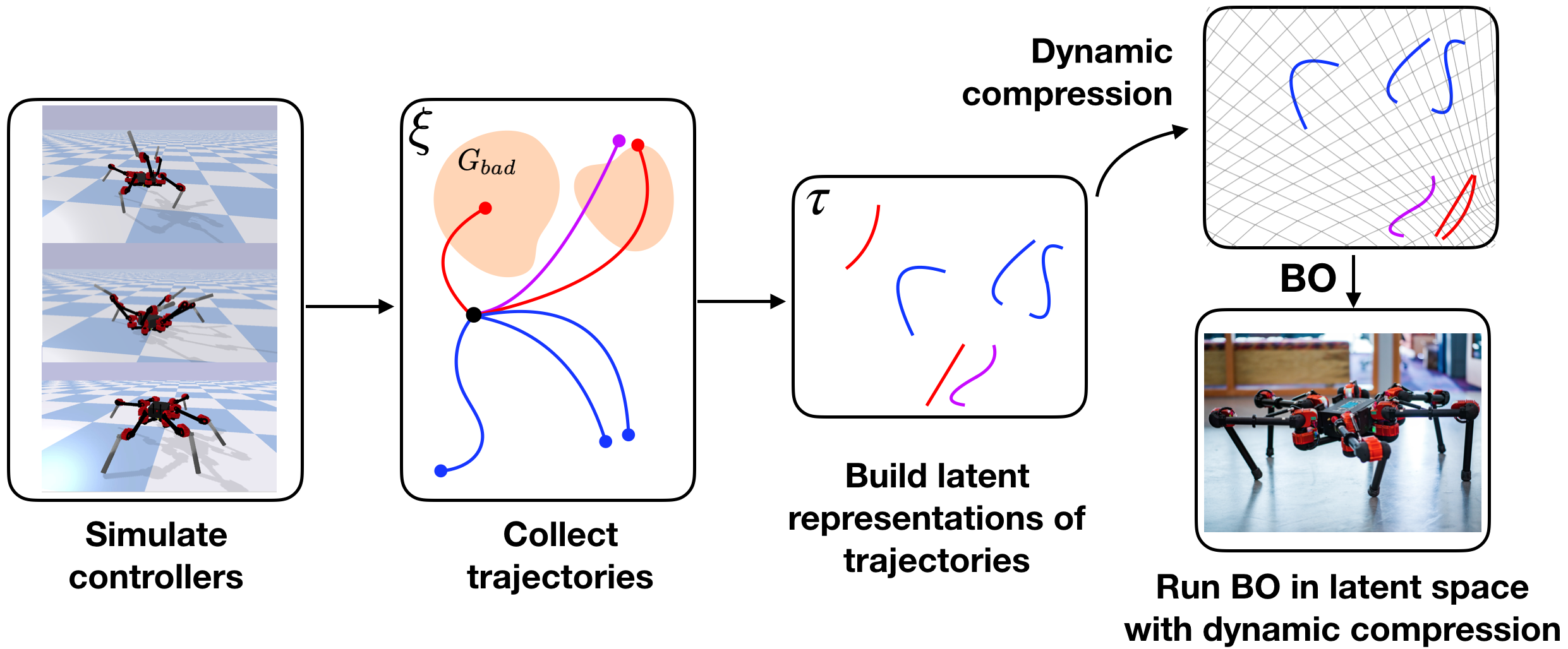}
    \vspace{2px}
    \caption{An overview of our approach: We start by simulating controllers and collecting their trajectories $\bxi$, along with the fraction of time spent in undesirable regions given by $G_{bad}$. Next, we learn to embed  trajectories into a lower-dimensional a space of latent paths $\btau$. We use dynamic compression to scale distances between latent paths based on their desirability. This dynamically compressed latent space is used for BO on hardware. \\
    Trajectory data $\bxi$ consists of high-frequency readings of robot joint angles and object position/velocity estimates (the framework can accommodate vision-based data in the future, but we do not experiment with it in this work).}
    \label{fig:approach_overview}
    \vspace{-13px}
\end{figure}

Global optimization in latent space can still suffer from sampling unsuccessful controllers, especially in the absence of dense rewards. One solution can be adding domain-specific constraints to point optimization in the right direction. While these can be hard to define in controller parameter space, frequently they can be easily expressed in observation/state space. For example, high velocities might be undesirable if they result in hard impacts. However, formulating this as constrained optimization could result in overly conservative controllers.
Instead, we incorporate controller desirability into BO by reducing exploration in the part of the trajectory space that leads to undesirable behavior. We compress the search space during BO dynamically by scaling the distance between controllers based on their desirability, initially inferred from simulation. BO can then quickly reject the undesirable parts of the search space, allowing for more exploration in the desirable parts. Figure~\ref{fig:approach_overview} gives an overview of the proposed approach.

We test our approach (SVAE-DC: informed kernel with Dynamic Compression) on a Daisy hexapod and an ABB Yumi manipulator on hardware\footnote{Video demonstrating hardware experiments: \url{https://youtu.be/2SvdwGZNrvY}}. We also conduct further simulation-based analysis on Daisy and two manipulators. On Daisy, our method consistently learns to walk in less than 10 hardware trials, outperforming uninformed BO. We also demonstrate significant gains on a nonprehensile manipulation task on Yumi. All latent components of our kernel can be adjusted online (by optimizing marginal likelihood as is done for BO hyperparameters). We anticipate that such adjustment could be useful for future works for settings with a medium budget of trials ($\approx$100+). Our code builds on the recently released BoTorch library~\cite{botorch} that supports highly scalable BO on GPUs. We open source our code for simulation environments, training and BO\footnote{SVAE-DC and BO code: \url{https://github.com/contactrika/bo-svae-dc}}.

\vspace{-3px}
\section{Background and Related Work}
\label{sec:background}
\vspace{-3px}
For learning with a small number of trials we turn to Bayesian Optimization (BO). BO can be thought of as a data-efficient RL method that obtains a reward only at the end of each trial/episode. For higher-dimensional robotics problems BO can benefit significantly from using simulation-based kernels. However, previous work required defining domain-specific features to be extracted from large-scale simulation data (see Section~\ref{sec:background_robotics}). 
Variational Autoencoders (VAEs)~\cite{kingma2013auto} provide an unsupervised alternative for embedding high-dimensional observations into a lower-dimensional space. For example, \cite{gomez2018automatic} recently used VAE in a Gaussian Process (GP) kernel to optimize chemical molecules. In robotics, VAEs have been used to process visual and tactile data (see~\cite{lesort2018state} for a survey).
We are interested in encoding trajectory data, so a sequential VAE (SVAE) could be applicable. \cite{yingzhen2018disentangled, fraccaro2017disentangled} show SVAEs learning latent dynamics. However, their physics simulations are low-dimensional (e.g. position of a 2D ball), sequences have length 20-30 steps, and the focus is on visual reconstruction. We aim to develop SVAE architecture that can easily handle simulations from full-scale robotics systems (state spaces 27D+) and much longer sequences (lengths 500-1000).

Our original motivation for embedding trajectory data into the kernel was Behavior Based Kernel (BBK)~\cite{wilson2014using}. On low-dimensional problems it matched the performance of PILCO~\cite{deisenroth2011pilco}, which is a popular data-efficient model-based RL algorithm for small domains. BBK is directly applicable only to stochastic policies, but we adapted it to our setting as BBK-KL baseline. We randomize simulator parameters when collecting trajectories. Hence even though the simulator and controllers are deterministic, each controller still induces a probability distribution over the trajectories. As proposed for BBK, for kernel distances we used symmetrized KL between trajectory distributions induced by the controllers. The generation and reconstruction parts of SVAE were used to estimate this KL. Since this baseline uses a neural network in the kernel, there is some relation to methods like~\cite{calandra2016manifold, wilson2016deep} (though these focused on GP regression, and did not incorporate trajectories).

\subsection{BO for Locomotion and Manipulation}
\label{sec:background_robotics}

Locomotion controllers most commonly used for real systems are structured and parametric~\cite{thatte2018method, feng2015optimization, gong2018feedback}. BO has been used to optimize their parameters, e.g.~\cite{calandra17thesis, lizotte2007automatic, tesch}. Typically, these methods take $\approx$40 trials for low-dimensional controllers (3-5D). For high-dimensional controllers further domain information is needed. For example \cite{cully2015robots} use simulation and user-defined features to transform the space of a 36-dimensional controller into 6D, making the search for walking controllers of a hexapod much more data-efficient. \cite{rai2019using} employ bipedal locomotion features to build informed kernels.

In manipulation, active learning and BO have been used, for example, for grasping~\cite{kroemer2010combining, montesano2012active}. These works did not incorporate simulation into the kernel, so their performance would be similar to BO with uninformed/standard kernel. \cite{antonova2018global} showed advantages of a simulation-based kernel, but needed grasping-specific features. Somewhat related are works in sim-to-real transfer, like~\cite{openai2018learning}, though many have visuomotor control as the focus (not considered here) and usually do not adapt online.
~\cite{chebotar2018closing} do adjust simulation parameters to match reality, so it would be interesting to combine this with BO in the future for global optimality (their work employs PPO, which is locally optimal).
Due to uncertainty over friction and contact forces, sim-to-real is challenging for non-prehensile problems. However, such motions can be useful to make solutions feasible (e.g pushing when the object is too large/heavy to lift or the goal is out of reach). \cite{peng2018sim, he2018zero, icra2019vpe} report success in transfer/adaptation on a push-to-goal task, showing the task is challenging but feasible. In our experiments we consider a `stable push' task: push two tall objects across a table without tipping them over. The further challenges come from interaction between objects and inability to recover from them tipping over.

\subsection{Challenges of Real-world Locomotion: the Need for Ultra Data-efficient Optimization}

Learning for legged locomotion can be a daunting task, since a robot needs to perfectly balance its interaction forces with the ground to move forward. For a hexapod robot, this means coordinating the movements of six legs, as well as the forces being applied on each leg. While it is easy to find a walking gait, it is extremely difficult to find a gait that can move forward at a reasonable speed. 

Recently, \cite{tan2018sim, li2018using} showed that RL can be used for locomotion on hardware. However, they learn conservative controllers in simulation and help transfer via system identification of actuator dynamics~\cite{tan2018sim} and a user-designed structured controller~\cite{li2018using}. While these methods can help, they do not guarantee that a controller learned in simulation will perform well on hardware.
\cite{minitaur} showed learning to walk on a Minitaur quadruped in only two hours. The Minitaur robot has 8 motors that control its longitudinal motion, and no actuation for lateral movements. In comparison, our hexapod (Daisy) has 18 motors, and has omni-directional movements. This makes the problem of controlling Daisy especially challenging, and would require significantly longer training. However, most present day locomotion robots get damaged from wear and tear when operated for long. For example, in the course of our experiments, we had to replace two motors, and fix issues such as faulty wiring and broken parts multiple times. With these considerations, we develop a ultra data-efficient approach that can learn controllers on Daisy in less than 10 hardware trials.

\vspace{-3px}
\section{SVAE-DC: Learning Informed Trajectory-based Embeddings}
\label{sec:svae_theory}
\vspace{-3px}
We model our setting as a joint Variational Inference problem: learning to compress/reconstruct trajectories while at the same time learning to associate controllers with their corresponding probability distributions over the latent paths. For this we develop a version of sequential VAE (SVAE). The training is guided by ELBO (Evidence Lower Bound) derived for our setting directly from the modeling assumptions and doesn't require any  auxiliary objectives.
First, we define notation:
\begin{itemize}[leftmargin=*]
\vspace{-2px}
\item[]$\pi_{\bx}$  :  policy/controller with parameters $\bx, \bx \in \R^D$; policies can be either deterministic or stochastic; for brevity we will refer to $\pi_{\bx}$ simply as `controller $\bx$'
\vspace{-1px}
\item[] $\bxi \equiv \bxi_{1:T}$  :  original trajectory for $T$ time steps containing high-frequency sensor readings
\vspace{-1px}
\item[] $\btau \equiv \btau_{1:K}$  :  latent space `path' (embedding of a trajectory)
\vspace{-1px}
\item[] $p(\bxi_{1:T} | \bx)$  :  a conditional probability distribution over the trajectories induced by controller $\bx$; the relationship between the controller and trajectories could be probabilistic either because the controller is stochastic, or because the simulator environment is stochastic, or both
\vspace{-1px}
\item[] $p(\btau_{1:K} | \bx)$  :  a conditional probability distribution over latent space paths induced controller by $\bx$
\item[] $G_{bad} : S \!\rightarrow\! \{0,1\}$ a map denoting whether an observation $\bxi_t \in S$ is within an undesirable region
\item[] $y$ : fraction of time $\bxi$ spends in undesirable regions; $\psi$ captures analogous notion in latent space
\end{itemize}
\vspace{-5px}

\begin{wrapfigure}{r}{0.26\textwidth}
\vspace{-15px}
\includegraphics[width=0.26\textwidth]{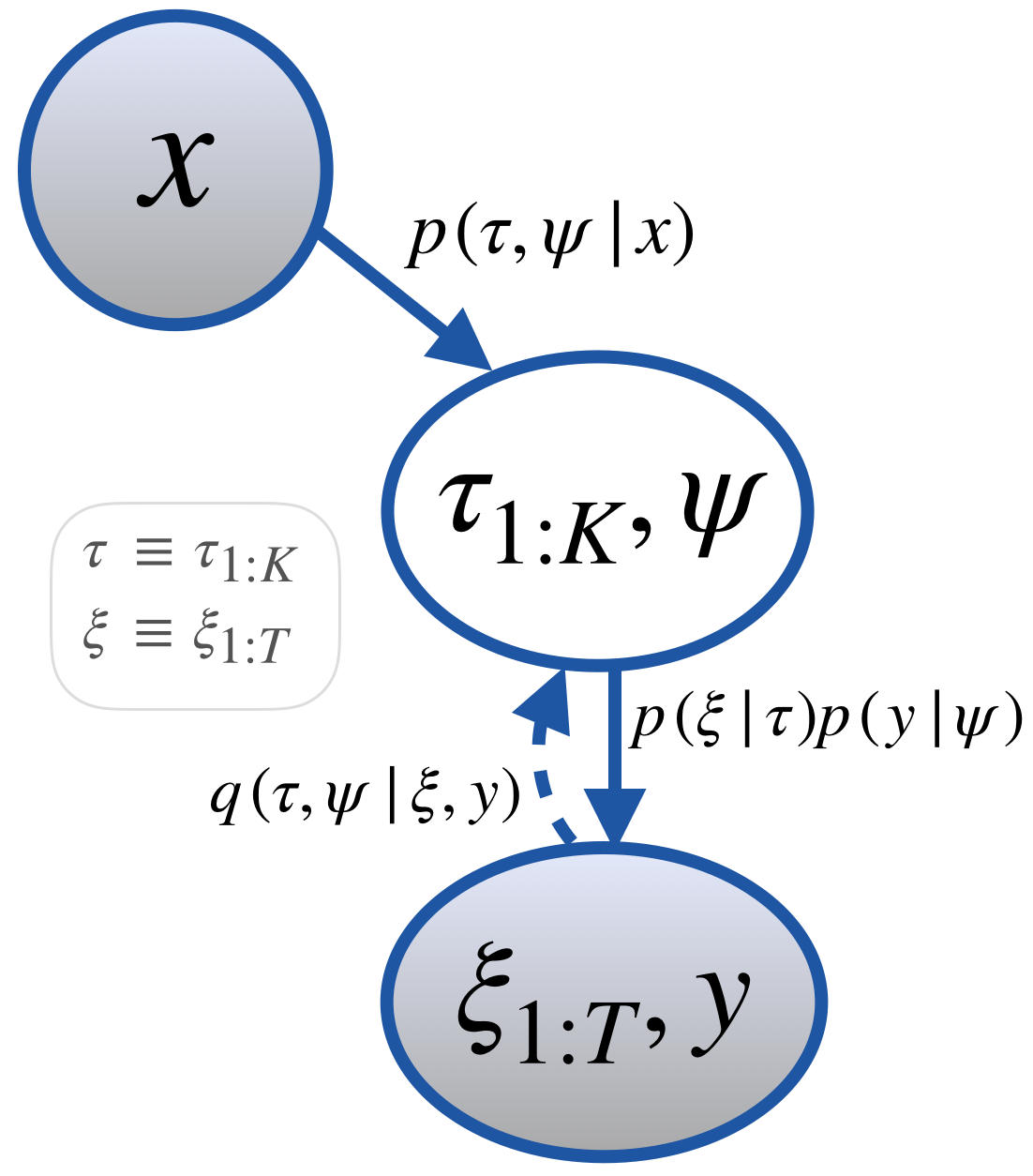}
\caption{A sketch of generative and inference model.}
\label{fig:model_dc}
\vspace{-15px}
\end{wrapfigure}
Our goal is to learn $p(\btau, \psi | \bx)$. $p(\btau | \bx)$ is analogous to $p(\bxi | \bx)$, only the paths are encoded in a lower-dimensional latent space. 
This is useful for constructing kernels for efficient BO on hardware.
As a measure of trajectory `quality' we can keep track of how long each trajectory spends in undesirable regions ($y$). For the latent paths we learn the analogous notion ($\psi$). We will not impose hard constraints during optimization, so $G_{bad}$ used to compute $y$ can be specified roughly with approximate guesses. Our framework also supports $G_{bad} : S \!\rightarrow\! [0,1]$, but for users it is frequently easier to make a rough thresholded estimate rather than providing smooth estimates or probabilities. The graphical model we construct for this setting is shown in Figure~\ref{fig:model_dc}. Not all independencies are captured by the illustration. So explicitly, the generative model is:
\\
$p(\btau, \psi, \bxi, y \ |\ \bx) = p(\btau_{1:K}, \psi | \bx) p(y | \psi) \prod_{t=1}^T p(\bxi_t | \bxi_{t-1}, \btau_{1:K}) $.
\vspace{3px}
\\
Approximate posterior is modeled by: $q(\btau, \psi, \bxi, y) = q(\btau_{1:K}, \psi | \bxi_{1:T}, y)$.\\
We collect trajectories $\bxi^{(i)}_{1:T}$ by simulating $N$ controllers with parameters $\bx^{(i)}$ for $T$ time steps.
We derive ELBO for this setting to maximize $\log p(Data) = \log p(\{\bx^{(i)}, \bxi_{1:T}^{(i)}\}_{i=1...N})$. Using `$\ \tilde{}\ $' over the variables to indicate samples from the current variational approximation, we get:
\vspace{-2px}
\begin{align}
\label{eq:elbo_dc_simple}
\begin{split}
\mathcal{L}^{DC}(\bw, \bphi | \bx, \bxi, y) = \mathbb{E}_{
\substack{\tilde{\btau},\tilde{\psi} \sim \\ q(\btau, \psi | \bxi, y )}} \Big[
\log p( \bxi | \tilde{\btau}) + \log p(y|\tilde{\psi}) + \log p(\tilde{\btau},\tilde{\psi} | \bx) 
- \log q(\tilde{\btau},\tilde{\psi} | \bxi, y) \Big]
\end{split}
\end{align}
Some aspects of this model resemble a setup from~\cite{louizos2016fair}; see derivation details in Appendix A.

\vspace{-3px}
\section{Bayesian Optimization with Dynamic Compression}
\label{sec:bo_theory}
\vspace{-3px}
In Bayesian Optimization (BO), the problem of optimizing controllers is viewed as finding controller parameters $\pmb{x}^*$ that optimize some objective function $f(\pmb{x})$:
$\displaystyle f(\pmb{x}^*) = \textstyle\max_{\pmb{x}} f(\pmb{x})$.
At each optimization trial BO optimizes an auxiliary function to select the next promising $\pmb{x}$ to evaluate. $f$ is commonly modeled with a Gaussian process (GP): $f(\pmb{x}) \sim \mathcal{GP}(m(\pmb{x}), k(\pmb{x}_i, \pmb{x}_j))$.

The key object is the kernel function $k(\cdot, \cdot)$, which encodes similarity between inputs. If $k(\pmb{x}_i, \pmb{x}_j)$ is large for inputs $\pmb{x}_i, \pmb{x}_j$, then $f(\pmb{x}_i$) strongly influences $f(\pmb{x}_j)$. One of the most widely used kernel functions is the Squared Exponential (SE) kernel:
$k_{SE}(\br \equiv |\pmb{x}_i \!\!-\!\! \pmb{x}_j|) = \sigma_k^2 \exp\big(\!-\!\tfrac{1}{2} \br^T \diag(\pmb{\ell})^{\!-\!2} \br \big)$, where $\sigma_k^2, \ \pmb{\ell}$ are signal variance and a vector of length scales respectively. $\sigma_k^2, \ \pmb{\ell}$ are called `hyperparameters' and are optimized automatically by maximizing marginal likelihood (\cite{BOtutorial2016}, Section~V-A).
SE belongs to a broader class of Mat\'ern kernels. One common parameter choice yields Mat\'ern$_{5/2}$:
$k_{{\text{Mat\'ern}}_{5/2}}(\br) = \big( 1 + \tfrac{\sqrt{5}\br}{\pmb{\ell}} + \tfrac{5\br^2}{3\pmb{\ell}^2} \big) \exp \big(\!-\!\tfrac{\sqrt{5}\br}{\pmb{\ell}}\big)$.
SE and Mat\'ern kernels are stationary, since they depend on $\br\!\equiv\!\pmb{x}_i\!-\!\pmb{x}_j \ \forall \pmb{x}_{i,j}$, and not on individual $\bx_i, \bx_j$. Section~\ref{sec:background_robotics} discussed recent work that showed how to effectively remove non-stationarity by using informed feature transforms for kernel computations. But these required extracting domain-specific features manually, or learning to fit a pre-defined set of features using a deterministic NN in a supervised way. 

We propose to use $p(\btau,\psi|\bx)$ learned by SVAE-DC. \cite{wilson2014using} showed that a `symmetrization' of KL divergence can be used to define a KL-based kernel for trajectories in the original space: 
\begin{equation}
k_{KL} = \exp(\text{-}\alpha D(\bx_i, \bx_j)) \ ; \ 
D(\bx_i, \bx_j) = \sqrt{KL(p(\bxi|\bx_i)||p(\bxi|\bx_j))} + \sqrt{KL(p(\bxi|\bx_j)||p(\bxi|\bx_i))}
\label{eq:BBK-KL}
\end{equation}
We could use this to define an analogous kernel in the latent space:
\eq{
k_{LKL} = \exp(\text{-}\alpha D_{\tau}(\bx_i, \bx_j)) ;
D_{\tau}(\bx_i, \bx_j) = \sqrt{KL(p(\btau|\bx_i)||p(\btau|\bx_j))} + \sqrt{KL(p(\btau|\bx_j)||p(\btau|\bx_i))}
}
In theory, this would be a natural way to define a path-based kernel in the latent space. However, it is widely known that Variational Inference tends to under-estimate variances in theory~\cite{minka2005divergence, bishop2006pattern} and in  practice~\cite{riquelme2018failure, tschiatschek2018variational}. This underestimation could negatively impact the practical performance of such kernel. Since we indeed observed variance under-estimation we implemented a version of the kernel to work with the latent means $\bar{\btau}_{\bx}, \bar{\psi}_{\bx} = E\big[ p(\btau,\psi|\bx) \big]$ directly.
We define our kernel function with:
\begin{align}
&\br_{\tau} = D_{\tau}(\bx_i, \bx_j) = \big| (1\!-\!\bar{y}_{\bx_i})\bar{{\btau}}_{\bx_i} -  (1\!-\!\bar{y}_{\bx_j}) \bar{\btau}_{\bx_j} \big| 
\ \ ; \quad y_{\bx} \sim p(y| \bar{\psi}_{\bx})
\label{eq:k_svae_dc_dist}
\end{align}
\begin{align}
&k_{SV\!AE\text{-}DC}(\bx_i, \bx_j) = \sigma_k^2 \exp\big(\!-\!\tfrac{1}{2} \br_{\tau}^T \diag(\pmb{\ell})^{\!-\!2} \br_{\tau} \big)
\label{eq:k_svae_dc}
\end{align}
This formulation is convenient in practice, since the form of Equation~\ref{eq:k_svae_dc} allows us to apply existing machinery for optimizing kernel hyperparameters $\sigma^2_k, \pmb{\ell}$. We can also define SVAE-DC-Mat\'ern version of the kernel by changing the form of Equation~\ref{eq:k_svae_dc} to the Mat\'ern function.
Scaling latent representations by $1\!-\!\bar{y}_{\bx}$ yields dynamic compression: latent representations that correspond to controllers frequently visiting undesirable parts of the space are scaled down. Hence `bad' controllers are brought closer together. This allows BO to reduce the number of samples from the `bad' parts of the space. `Dynamic compression' here means this search space transformation is applied after SVAE training, in addition to the compression obtained by SVAE's dimensionality reduction.
The scaling can be made non-linear with $sigmoid(\alpha(\bar{y}_{\bx}-c))$. This can help achieving aggressive compression in settings like ours with an extremely small budget of trials. The additional parameters $\alpha, c$, as well as $p(\btau,\psi|\bx), p(y|\psi)$ can be optimized online in the same way as BO hyperparameters.

Overall, SVAE-DC and the resulting kernel described above allow us to obtain a fully automatic way of learning latent trajectory embeddings in unsupervised way. For domains where $G_{bad}$ is given we can also achieve dynamic compression of the latent space, making BO ultra data-efficient. All the components used during BO can be optimized online via the same methods already implemented for automatically adjusting BO hyperparameters. 

\vspace{-3px}
\section{SVAE-DC: NN Architectures and Training}
\vspace{-3px}
Guided by prior literature we experimented with RNNs, LSTMs, and sequence-to-sequence RNNs. Learning was slow and frequently unsuccessful, despite trying adaptive learning rates, manual tuning, weighting various parts of ELBO. Using MLPs instead did not improve performance either. \cite{BaiTCN2018} notes that CNNs can succeed on sequence data, but one recent alternative (Quasi-RNNs~\cite{bradbury2016quasi}) did not yield a notable improvement for us. Instead, an effective idea we had was to view dimensions of $\bxi_t, \btau_k$ as different channels. Then we could feed $\bxi_{1:T}$ to 1D convolutional layers to learn $q(\btau|\bxi)$, de-convolutional for $p(\bxi|\btau)$. With that, for all our experiments (all different robot and controller architectures) we were able to use the same network parameters: 3-layer 1D convolutions with [32, 64, 128] channels (reverse order for de-convolutions; kernel size 4, stride 2) followed by MLP layer for $\mu, \sigma$ outputs. We were also able to use same latent space sizes: 3-dimensional $\btau$, latent sequence length $K\text{=}3$. This yielded a small 9D optimization space for BO, which is highly desirable for optimization with few trials. Notably, this NN architecture also retained good reconstruction accuracy, not far from results with larger latent spaces ($\btau\text{=}6D,12D; K\text{=}5,15$) and hidden sizes (256-1024). We also interpreted $\bx$ as a sequence of length 1 and used de-convolutional architecture for $p(\btau|\bx)$. It had 4-layers with [512, 256, 128, 128] channels, since $p(\btau|\bx)$ was one of the key parts for BO (though a smaller CNN or MLP could have sufficed). For $p(y|\psi)$ we used a 2-layer MLP (hidden size 64). Training took $\approx$30-180 minutes on 1 GPU, using $1e\text{-}4$ learning rate (halving after each 5K gradient updates, stopping at $1e\text{-}5$). See Appendix B for reconstruction/generation visualizations.

\section{Locomotion on the Daisy Hexapod}

\begin{wrapfigure}{r}{0.40\textwidth}
    \vspace{-12px}
    \centering
    \includegraphics[width=0.40\textwidth]{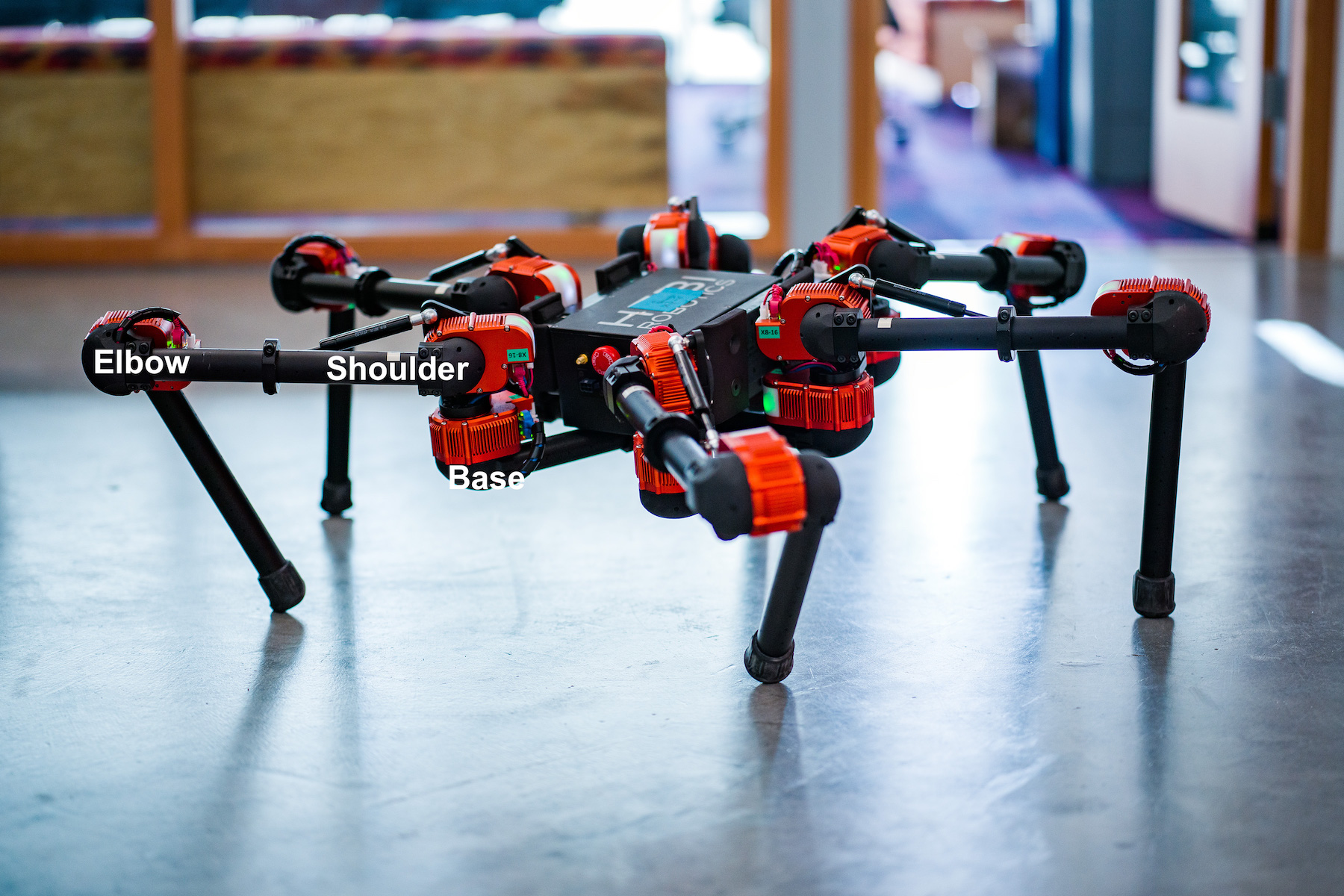}
    \caption{\small{Daisy hexapod used in this work.}}
    \label{fig:daisy_robot}
    \vspace{-10px}
\end{wrapfigure}

For our locomotion experiments we used Daisy robot (Figure~\ref{fig:daisy_robot}) from Hebi robotics \cite{hebi}. It has six legs, each with 3 motors -- base, shoulder and elbow.
The robot is practically omni-directional, however, the motors are velocity limited, so the robot is unable to achieve very high velocities. Vive tracking system was used to measure robot's position in the global frame for rewards.

In general, locomotion is a hard learning problem, but complex high degree-of-freedom robots further complicate it. While in simulation all 6 legs of Daisy are identical, each motor has a slightly different behavior on hardware. This also depends on the environment, and makes it extremely hard to predict the robot's behavior from simulation. For example, one of our successful straight-walking controllers from simulation, turns left when executed on a carpet floor, but turns right on a wooden floor.
This raises the need for learning approaches that can transfer information from simulation to hardware, without suffering too much from the mismatch between the two.
We simulated the Daisy robot in PyBullet~\cite{pybullet}.
The simulator was fast, but did not have an accurate contact model with the ground. While free-space motion of individual joints transferred to hardware, the overall behavior of the robot when interacting with the ground was very different between simulation and hardware. As a result, rewards obtained by controllers in simulation could be significantly different on hardware.

\textbf{Daisy Controllers}: We used Central Pattern Generators (CPGs) from \cite{crespi2008online}.
These are capable of generating a large number of locomotion gaits by changing the frequency, amplitude, and offset of each joint, as well as the relative phase differences between joints. Different CPG parameters can be restricted to obtain controllers with various dimensionalities. We experimented with 11D controller on hardware and 27D in simulation. 
For hardware, we assume that all joints have the same amplitude, frequency and offset (3 parameters), all base motors have independent phases (6 parameters), all shoulders and elbows have the same phase difference w.r.t. the base (2 parameters).
This assumption implies that all joints are treated identically, which doesn't always hold, since each motor has slightly different tracking and bandwidth. In the future, we would like to use alternatives that allow each motor to learn independently. For simulation: base, shoulder and elbow joints were allowed to have independent amplitudes, frequencies and offsets, but fixed across the six legs (9 parameters); each of the 18 joints was allowed to have an independent phase (18 parameters).

\vspace{-10px}
\subsection{Daisy Experiments}
\vspace{-3px}
For SVAE-DC training we sampled $500,000$ controllers randomly in simulation and collected the corresponding trajectories for $1000$ time steps ($\approx\!16.5s$). For dynamic compression the states were marked as undesirable if they had: high joint velocities (more than 10rad/sec); robot base tilting by more than 60\degree in roll and pitch, elbows hitting the ground; height of the base outside of [0.1, 0.7]cm from the ground. These aimed to reduce the chance of robot breaking: controllers with high joint velocities can harm the motors on impact with the ground; tilting the torso can cause the robot to fall on its back; scraping the ground or lifting off and then falling can cause further damage. Since our BO trials were in a narrow walkway, we also marked as undesirable states deviating more than $0.5$m from the starting $x$-coordinate of the base.
The objective function for BO was: $f(\bx) = 10 y_{final} \!-\! N_{high\_vel}$, where $y_{final}$ was the final $y$-coordinate of the robot (how much the robot walked forward), $N_{high\_vel}$ was the number of timesteps with velocities exceeding 10rad/sec. All BO experiments used UCB acquisition function (with $\beta\!=\!1$).

\begin{wrapfigure}{r}{0.35\textwidth}
    \centering
    \vspace{-12px}
    \includegraphics[width=0.35\textwidth]{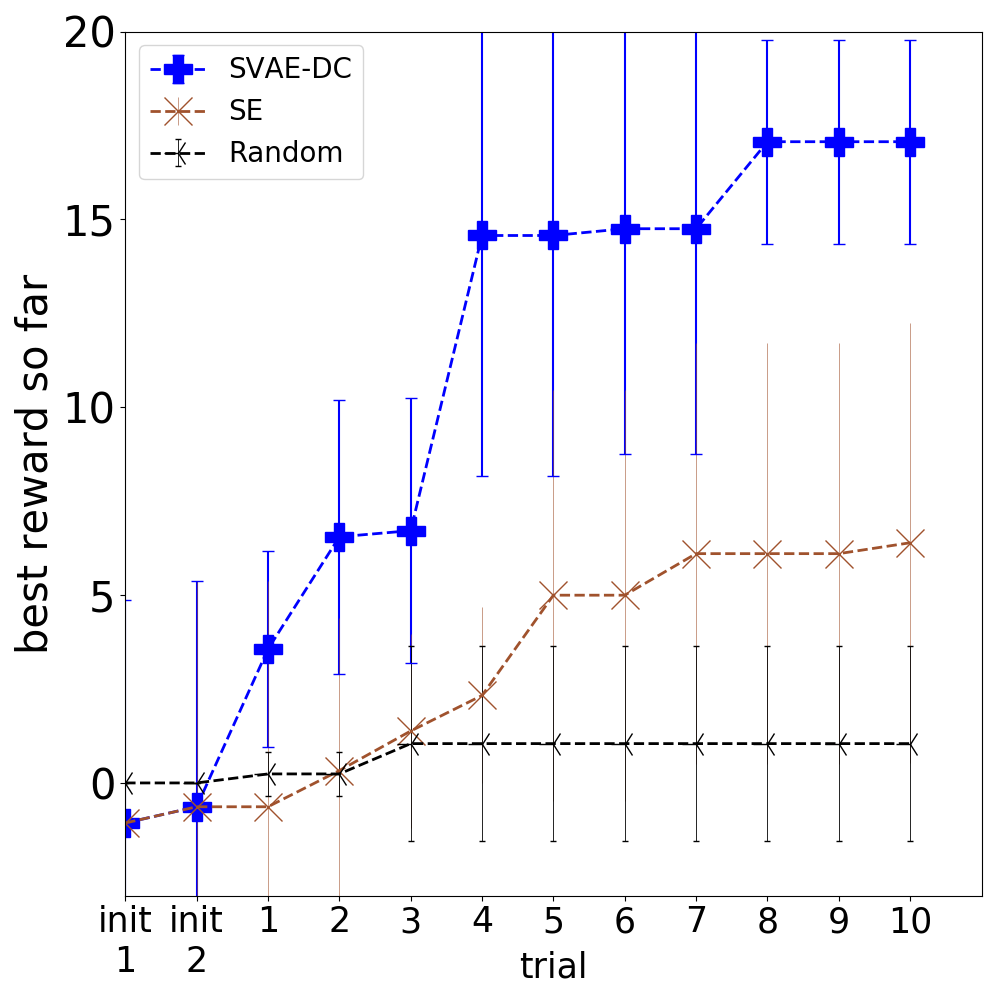}
    \caption{\small{BO on Daisy hardware. Mean over 5 runs, 90\% CIs.}}
    \vspace{-12px}
    \label{fig:daisy_hw}
\end{wrapfigure}
We completed 5 runs of BO on the Daisy robot hardware, initializing with 2 random samples, followed by 10 trials of BO (Figure~\ref{fig:daisy_hw}). BO with SE kernel used the same initialization as BO with SVAE-DC kernel.
For Daisy robot on hardware the controller would be considered acceptable if it walked forward for more than $1.5m$ during a trial of 25 seconds. For comparison to random search we sampled 60 controllers at random. Of these only 2 were able to walk forward a distance of over $1.5m$ in $25s$. So the problem was challenging, as the chance of randomly sampling a successful controller was $<\!\!4$\%. BO with SVAE-DC kernel found walking controllers reliably in all 5/5 runs within fewer than 10 trials. In contrast, both BO with SE found forward walking controllers only in 2/5 runs.

For simulation experiments, we created an artificial `sim-to-real' gap, allowing to gauge the potential for simulation-based kernels without running all the experiments on hardware. For each BO run we randomly sampled ground restitution parameters, and kept them fixed for all trials within a run. Hence simulation-based kernels did not have full information about the exact properties of the environment used during BO (even though the range of parameters was the same as for data collection). Kernels were informed about performance on a range of parameters, but could have caused negative transfer by lagging to identify controllers that perform best in a particular setting (not only well on average across settings).

\begin{wrapfigure}{r}{0.33\textwidth}
    \centering
    \vspace{-5px}
    \includegraphics[width=0.33\textwidth]{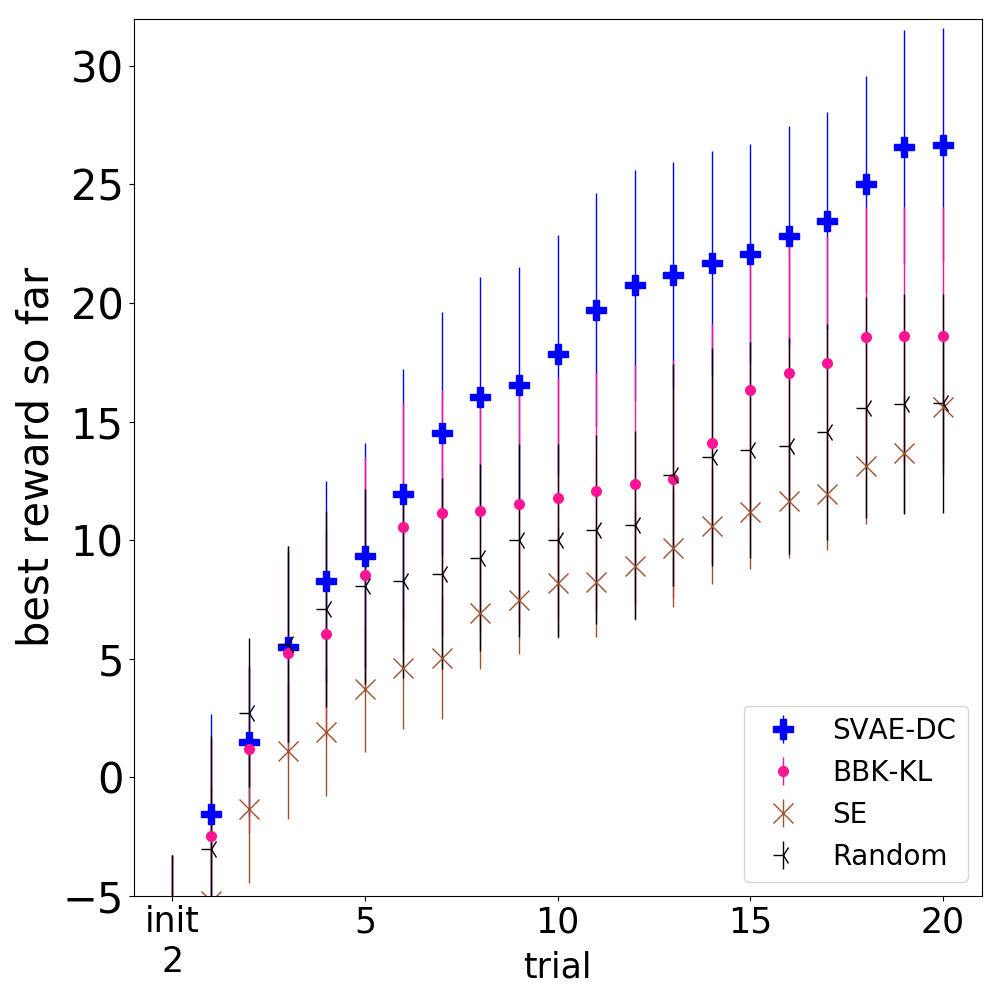}
    \caption{BO for Daisy in simulation. Means over 50 runs, 90\% CIs.}
    \label{fig:daisy_sim}
    \vspace{-20px}
\end{wrapfigure}
Figure~\ref{fig:daisy_sim} shows BO with 27D controller. BO with SVAE-DC outperformed all baselines. BBK-KL kernel obtained smaller improvements over SE and Random baselines. This indicated that a trajectory-based kernel was useful even when optimizing a high-dimensional controller, although BBK-KL benefits were greatly diminished compared to BBK results for 2-4 dimensional controllers reported in prior work. In these experiments, SVAE without dynamic compression was very similar to SE (omitted from the plot for clarity, since it was overlapping with SE). This showed that dimensionality reduction alone does not guarantee improvement (even when the latent space contains information needed to decode back into the space of original trajectories).

\section{Manipulation Experiments}

\begin{wrapfigure}{r}{0.35\textwidth}
    \vspace{-15px}
    \centering
    \includegraphics[width=0.35\textwidth]{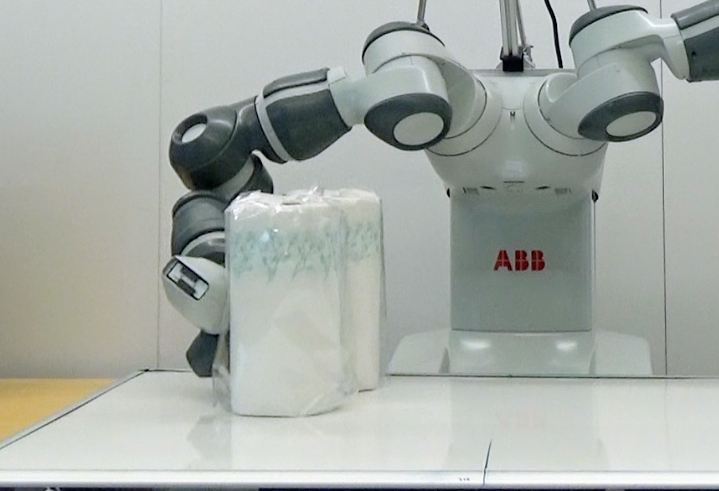}
    \caption{\small{``Stable push'' task with Yumi}}
    \label{fig:yumi_push_task}
    \vspace{-7px}
\end{wrapfigure}
Our manipulation task was to push two objects from one side of the table to another without tipping them over. For Yumi environment the objects had mass and inertial properties similar to paper towel rolls (mass of 150g, 22cm height, 5cm radius); for Franka these had properties similar to wooden rolls (2kg, 22cm height, 8cm radius).
Compared to `push-to-target' task, our task had two different challenges. The objects were likely to come into contact with each other (not only the robot arm). Moreover, they could easily tip over, especially if forces were applied above an object's center of mass. Reward was given only at the end of the task: the distance each upright object moved in the desired direction minus a penalty for objects that tipped over (with $y_{max}$ being table width): $f(\bx) =\!\! \textstyle\sum_{i} \big[ (y_{final}^{obj_i} \!-\! y_{start}^{obj_i})\mathbbm{1}_{obj_i \in Up} \!- y_{max} \mathbbm{1}_{obj_i  \in Tipped}\big]$.

\textbf{Controllers}: We tested our approach on two types of controllers: 1) joint velocity controller suitable for robots like ABB Yumi and 2) torque controller suitable for robots like Franka Emika. The first was parameterized by 6 joint velocity ``waypoints'', one target velocity for each joint of the robot arm (so $6\!\cdot\!7\!=\!\!42$ parameters for a 7DoF arm). Each ``waypoint'' also had a duration parameter that specified the fraction of time to be spent attaining the desired joint velocities. Overall this yielded a 48-dimensional parametric controller.
The second controller type was aimed to be safe to use on robots with torque control that are more powerful than ABB Yumi. Instead of exploring randomly in torque space, we designed a parametric controller with desired waypoints in end-effector space. Each of the 6 waypoints had 6 parameters for the pose (3D position, 3D orientation) and 2 parameters for controller proportional and derivative gains. Overall this yielded a 48-dimensional parametric controller: $6\cdot\!(6\!+\!2)$.
This controller interpolated between the waypoints using a $5^{th}$ order minimum jerk trajectory for positions, and used linear interpolation for orientations. End effector Jacobian for the corresponding robot model was used to convert to joint torques.

\subsection{Experimental Setup and Results}

For training SVAE-DC we collected 500,000 simulated trajectories for both Yumi and Franka robot. These contained joint angles of the robot and object poses at each time step (1000 steps for Yumi and 500 steps for Franka, simulated with pybullet at 500Hz).
A step on the trajectory was marked as undesirable $\big(G_{bad}(\pmb{\xi}_t)\!=\!1\big)$ when: any object tipped over or was pushed beyond the table; robot collided with the table or the end effector was outside of main workspace (not over the table area). Mass, friction and restitution of the objects were randomized at the start of each episode/trajectory. Randomization ranges were set to roughly resemble variability of how real-world objects behaved.

ABB Yumi robot available to us could operate effectively only at low velocities ($\tfrac{1}{5}$ of simulation maximum). High-velocity trajectories successful in simulation yielded different results on hardware. To prevent Yumi from shutting down due to high load we stopped execution if the robot's arm extended too far outside the main workspace, also stopped if it was about to collide with the table (giving $-2 y_{max}$ reward in such cases). These factors caused a large sim-real gap. Nonetheless,

\begin{wrapfigure}{r}{0.35\textwidth}
    \vspace{-5px}
    \centering
    \includegraphics[width=0.35\textwidth]{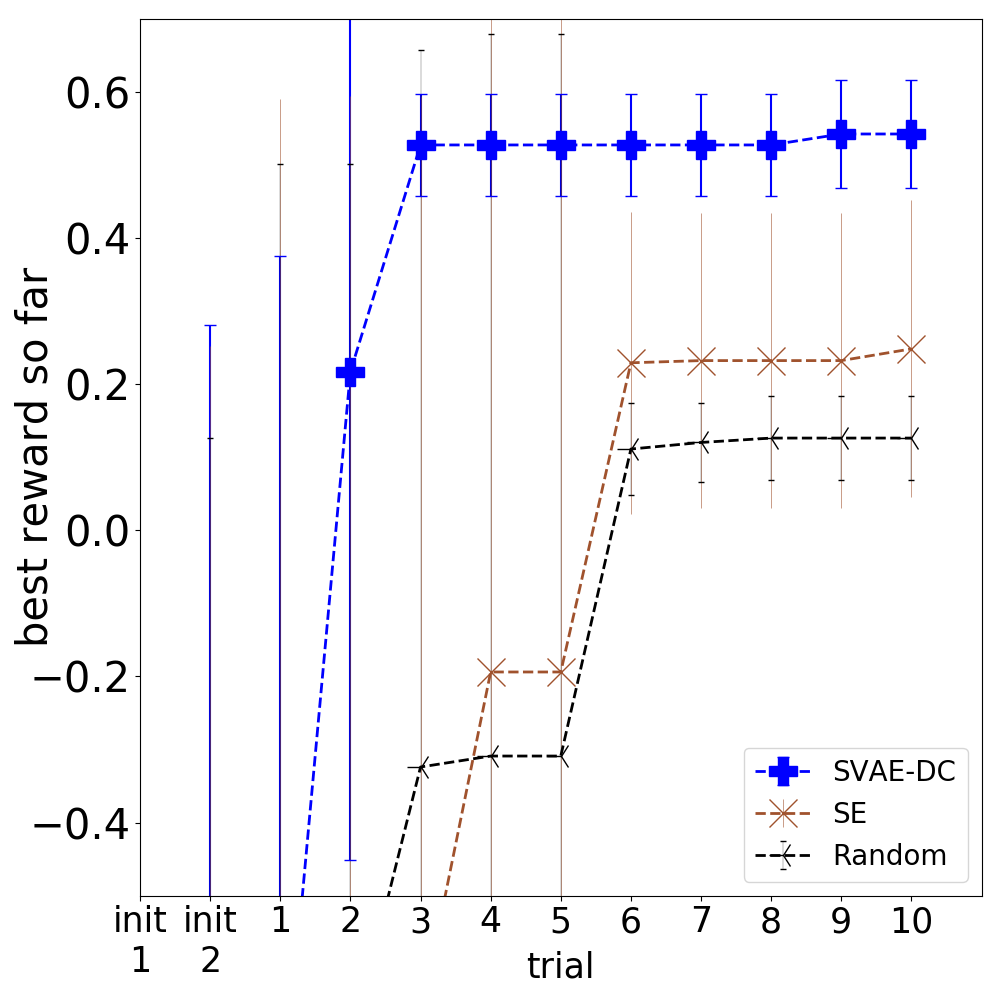}
    \caption{BO on ABB Yumi hardware (mean of 5 runs, 90\% CIs).}
    \label{fig:yumi_bo_hw}
    \vspace{-15px}
\end{wrapfigure}
BO with SVAE-DC kernel was still able to significantly outperform BO with SE (Figure~\ref{fig:yumi_bo_hw}). Even when controllers successful in simulation yielded very different outcomes on hardware, SVAE-DC kernel was still able to find well-performing alternatives (more conservative, yet successful on hardware).

\begin{wrapfigure}{r}{0.35\textwidth}
    \vspace{-10px}
    \centering
    \includegraphics[width=0.35\textwidth]{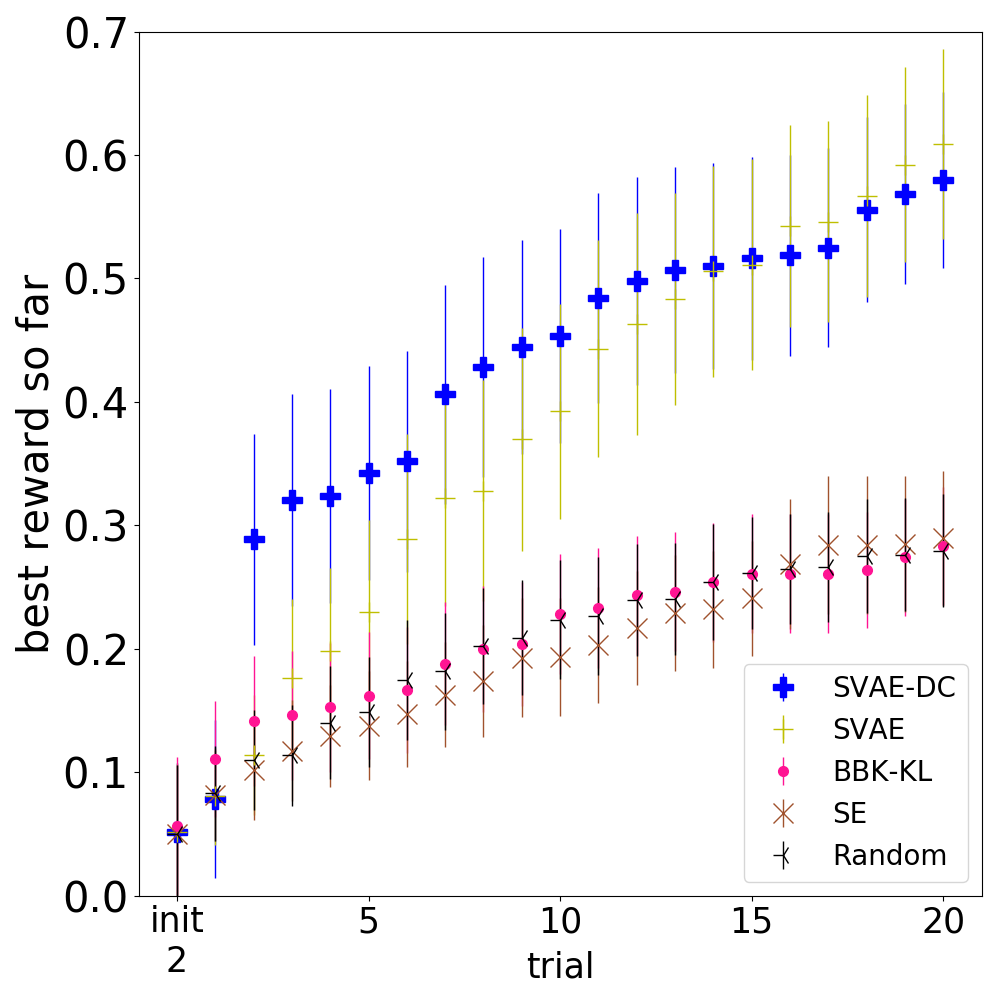}
    \caption{BO on ABB Yumi simulation (mean of 50 runs, 90\% CIs).}
    \label{fig:yumi_bo_sim}
    \vspace{-10px}
\end{wrapfigure}

\begin{figure}[t!]
    \centering
    \includegraphics[width=0.325\textwidth]{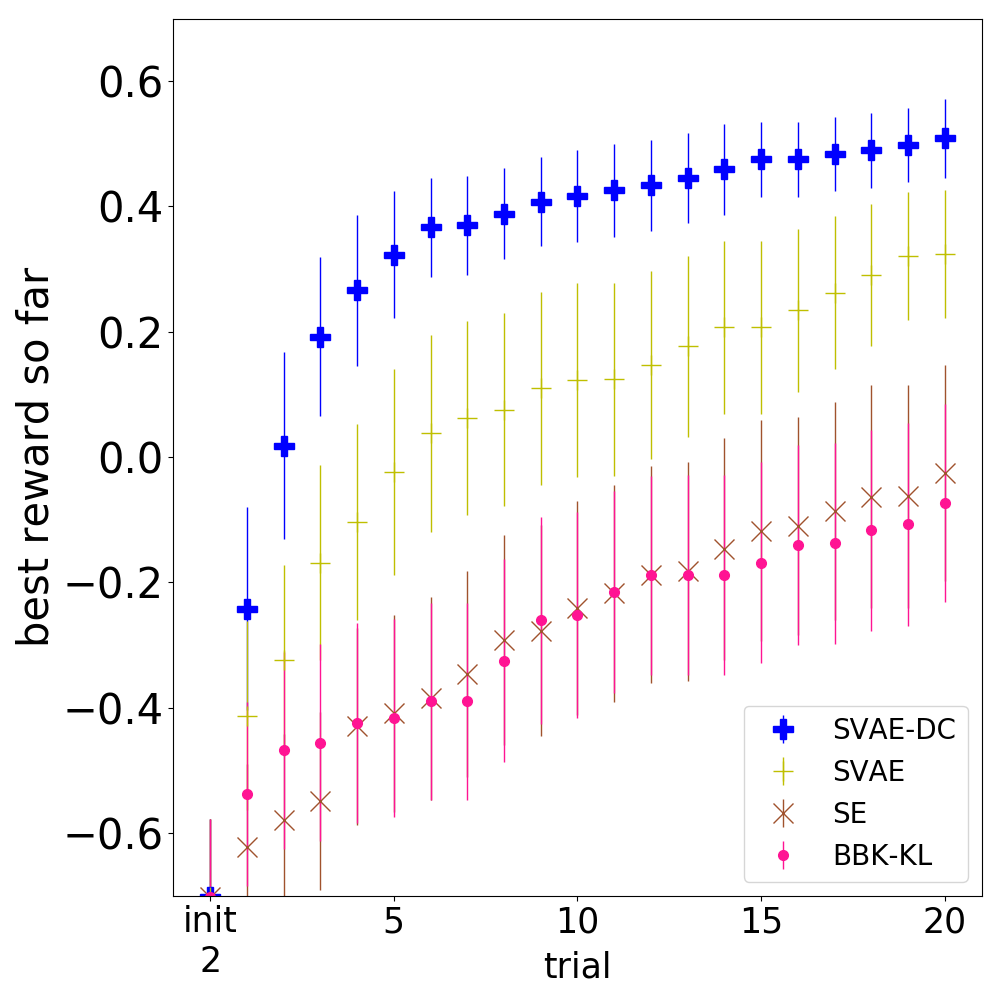}
    \includegraphics[width=0.325\textwidth]{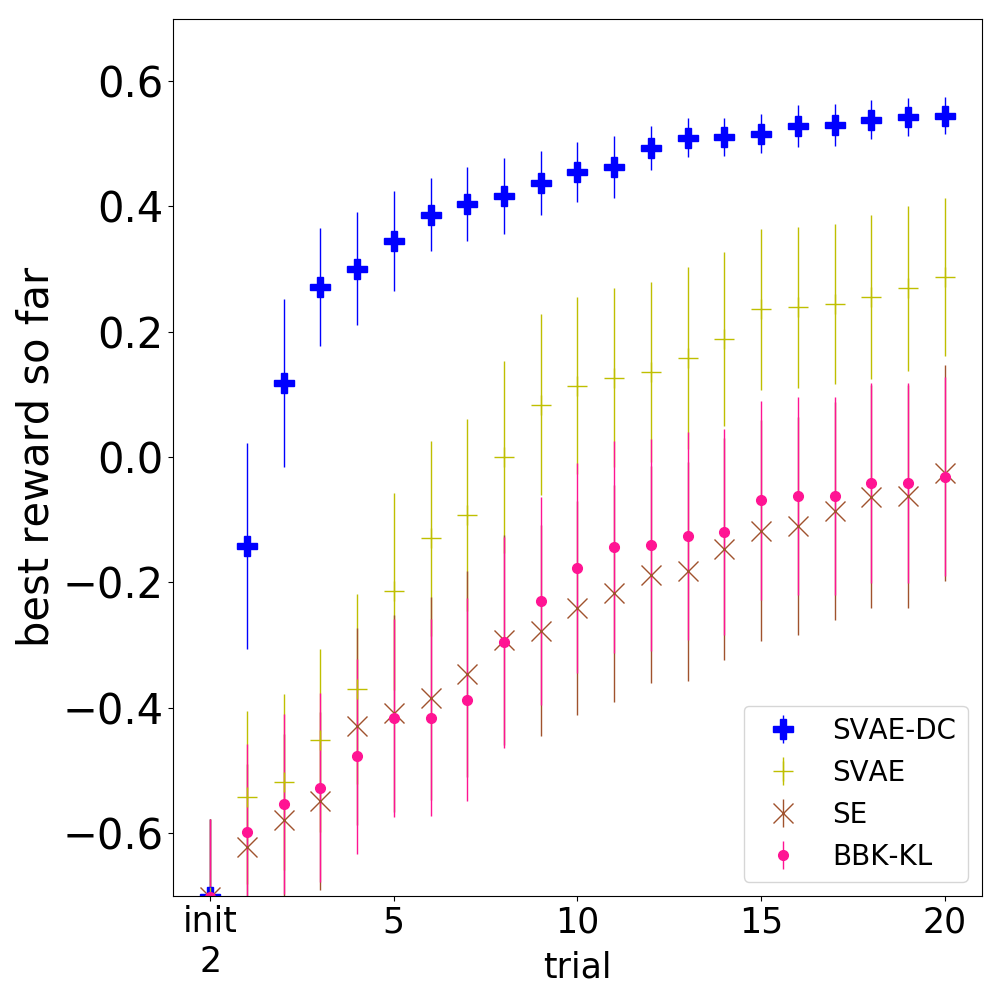}
    \includegraphics[width=0.325\textwidth]{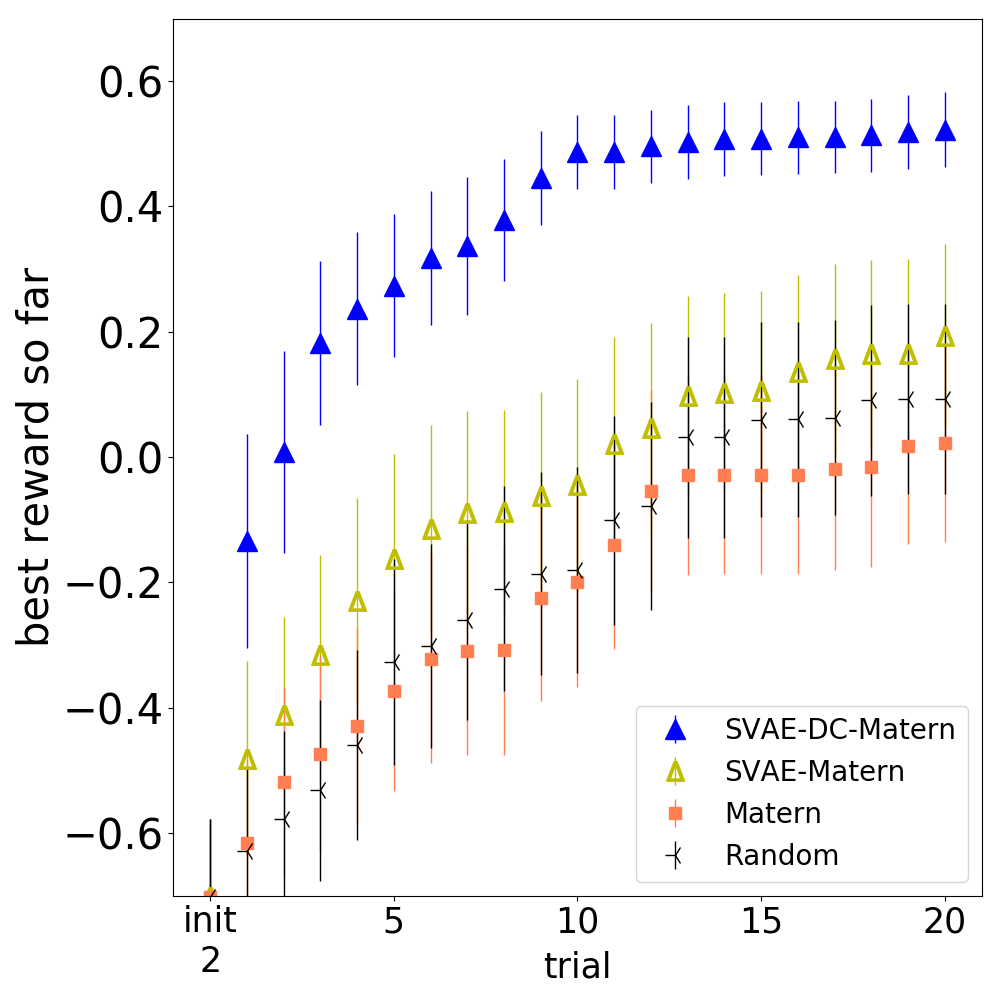}
    \caption{BO with various kernels on Franka Emika simulation. Left: SVAE trained with same parameters as in all the previous experiments. Middle: SVAE with larger latent space and NNs. Right: Matern used as outer function for all kernel. The plots show means over 50 runs, 90\% CIs.}
    \label{fig:franka_bo}
    \vspace{-15px}
\end{figure}

For simulation experiments with manipulators we emulated `sim-to-real' gap as with Daisy simulation: sampled different object properties (mass, friction, restitution) at the start of each BO run. Results in Figure~\ref{fig:yumi_bo_sim} show that BO on Yumi with SVAE-DC kernel yielded substantial improvement over all baselines. BO in the latent space of SVAE (without dynamic compression) was also able to substantially outperform all baselines, matching SVAE-DC gains after $\approx$15 trials. \\
Figure~\ref{fig:franka_bo} shows BO results on Franka Emika simulation (left).
\vspace{-5px}

Furthermore, we analyze how increasing the size of SVAE latent space and NNs impacts performance (middle). The larger latent space is $6\!\cdot\!5\!\!=\!\!30$D (vs 9D in other experiments), the hidden layer size of NNs is increased from 128 to 256.
Larger latent space implies larger search space for BO, which could impair data efficiency. Indeed, we see what BO with SVAE kernel outperforms BBK-KL and SE kernels not as early as before. However, BO with SVAE-DC is able to keep the gains and even decrease the variance between runs (well-performing points are found more reliably). This indicates that dynamic compression could counter-balance increase in kernel dimensionality. Finally, we experimented with Mat\'ern kernel (right plot in Figure~\ref{fig:franka_bo}), but it did not show benefits over using SE kernel. We attempted changing hyperparameter prior and restricting hyperparameter ranges, but it did not consistently outperform random search (same held for SE in high dimensions). The performance of BO with SVAE kernel using Mat\'ern as outer kernel function showed modest improvement over baselines. In contrast, BO with SVAE-DC kernel kept most improvements.

\vspace{-10px}
\section{Conclusion}
\vspace{-10px}
In this work employed BO to optimize robot controllers with a small budget of trials. Previously, the success of BO has been either limited to low-dimensional controllers or required simulation-based kernels with domain-specific features. We proposed an unsupervised alternative with sequential variational autoencoder. We used it to embed simulated trajectories into a latent space, and to jointly learn relating controllers with latent space paths they induce. Furthermore, we provided a mechanism for dynamic compression, helping BO reject undesirable regions quickly, and explore more in other regions. Our approach yielded ultra-data efficient BO in hardware experiments with hexapod locomotion and a manipulation task, using the same SVAE-DC architecture, training and BO parameters.

\clearpage
\acknowledgments{\thanks{This research was supported in part by the Knut and Alice Wallenberg Foundation.}}
\bibliography{ms}  

\section*{Appendix A: SVAE-DC Modeling Details}

The backbone of our model is inspired by hierarchical constructions, like those developed in~\cite{louizos2016fair, kingma2014semi}. However, these works considered supervised and semi-supervised settings, where a discrete label was associated with each high-dimensional data point (e.g. a label for an image). We are dealing with sequential trajectory data instead, so the internal structure of our data is different. But for the moment let us think about each trajectory as a point in some high-dimensional space. Our idea is to interpret controllers $\bx$ as continuous `labels' for trajectories $\bxi$. Then, on a high level, for random variables $\bx, \bxi, \btau$ we can extend the standard ELBO bound as follows:
\begin{align}
\label{eq:elbo_simple}
\mathcal{L}(\bw, \bphi | \bxi, \bx) &= \mathbb{E}_{
\tilde{\btau} \sim q_{\phi_\tau}(\btau | \bxi )} \Big[ \log p_{w_{\xi}}( \bxi | \tilde{\btau})
- \log q_{\phi_{\tau}}(\tilde{\btau} | \bxi)
+ \log p_{w_{\tau}}(\tilde{\btau} | \bx) \Big]
\end{align}

\begin{wrapfigure}{r}{0.25\textwidth}
\vspace{-10px}
\includegraphics[width=0.25\textwidth]{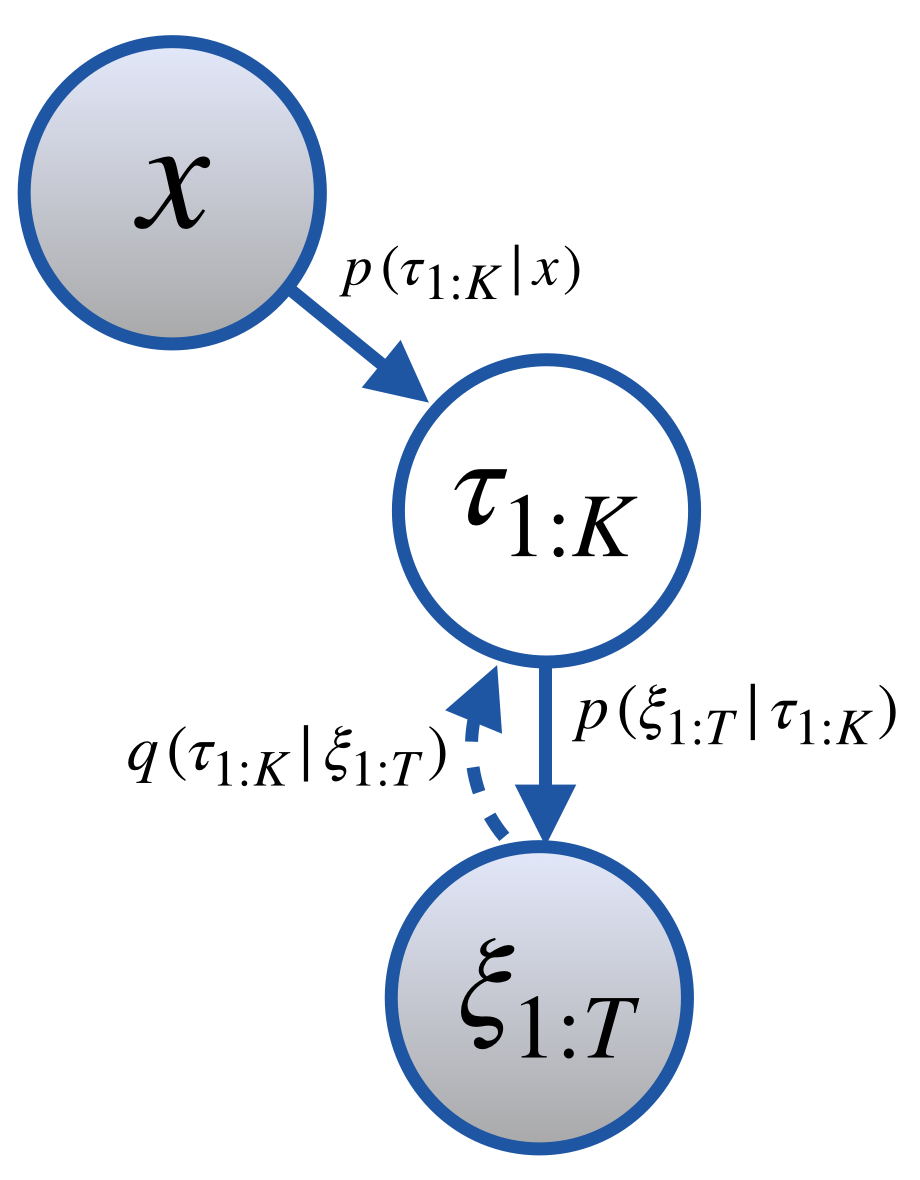}
\caption{Backbone of our SVAE generative model and inference}
\vspace{-50px}
\label{fig:model_simple}
\end{wrapfigure}
In the above, $\bphi = [ \bphi_\xi ]$ denote parameters of the variational approximation, $\bw = [\bw_{\tau}, \bw_{\xi}]$ denote the parameters of the generative part of the model. In our work, $\bphi, \bw$ are weights of deep neural networks. It is customary to drop subscripts indicating NN weight parameters and write $q,p$ for a shorthand notation.

The derivation for the above is similar to \cite{louizos2016fair}. We can also recognize the similarity to a standard ELBO for a simplified model without $\bx$:
\eq{
\mathcal{L}(q,p) &= \int q(\btau | \bxi) \log \frac{p(\btau,\bxi)}{q(\btau | \bxi)} \ d\btau \\
&= \mathbb{E}_{q(\btau | \bxi)} \Big[ \log p(\bxi | \btau) \Big] - KL(q(\btau | \bxi) || p(\btau)) \\
&= \mathbb{E}_{q(\btau | \bxi)} \Big[ \log p(\bxi | \btau)  - q(\btau | \bxi) + p(\btau) \Big]
}
In our case, $\bx$ is observed (we know which controller is executed when we obtain a trajectory $\bxi$), so there is no further uncertainty about $\bx$. Also, from the independencies in the model we see that the rest of the variables are independent of $\bx$ given $\btau$. Hence $p(\btau|\bx)$ conditioning is the only modification that appears in $\mathcal{L}(\bw, \bphi | \bxi, \bx)$. This is why terms like $p(\bx), q(\bx|\btau)$ do not appear in our ELBO, but they would have been included if we also had a non-trivial prior $p(\bx)$. Our construction treats $\bx$ and $\bxi$ as observed data, which is in fact what we have available from simulating trajectories. For data collection we can sample controllers at random, since we assume access to a relatively inexpensive simulator (in a sense that it is viable to simulate 100K+ trajectories for training). So we don't need a sophisticated prior for $p(\bx)$.

To derive the full SVAE-DC ELBO we can use the decomposition assumptions of the generative model and the approximate posterior:
\eq{
\text{Generative model: }& p(\btau, \psi, \bxi, y \ |\ \bx) = p(\btau_{1:K}, \psi | \bx) p(y | \psi)  \textstyle \prod_{t=1}^T p(\bxi_t | \bxi_{t-1}, \btau_{1:K})
\\
\text{Approximate posterior: }& q(\btau, \psi, \bxi, y) = q(\btau_{1:K}, \psi | \bxi_{1:T}, y)
}
\eq{
\mathcal{L}^{DC}(\bw, \bphi | \bx, \bxi, y) &= 
\mathbb{E}_{
\tilde{\btau}, \tilde{\psi} \sim q(\btau, \psi | \bxi, y )} \Big[ \log p( \bxi, y | \tilde{\btau}, \psi)
- \log q(\tilde{\btau}, \psi | \bxi, y)
+ \log p(\tilde{\btau}, \psi | \bx) \Big]
}
\begin{align}
\label{eq:elbo_svae_dc}
\mathcal{L}^{DC}(\bw, \bphi | \bx, \bxi, y) &= \mathbb{E}_{
\substack{\tilde{\btau},\tilde{\psi} \sim \\ q(\btau, \psi | \bxi, y )}} \Big[
\log p( \bxi | \tilde{\btau}) + \log p(y|\tilde{\psi}) + \log p(\tilde{\btau},\tilde{\psi} | \bx) 
- \log q(\tilde{\btau},\tilde{\psi} | \bxi, y) \Big]
\end{align}
The 4 terms inside the expectation in Equation~\ref{eq:elbo_svae_dc} above are the 4 neural networks whose parameters will be optimized by gradient ascent to maximize $\mathcal{L}^{DC}$. The choices for their architectures are described in the main paper. For ease of implementation we treat the outputs of $p(\btau, \psi | \bx), q(\btau, \psi | \bxi, y)$ as a single latent code $[\btau, \psi]$, and separate it into components only when needed (e.g. to feed only the $\psi$ part into $p(y|\psi)$ NN, etc.

One advantage of our formulation is that it is agnostic to whether policies/controllers are stochastic or deterministic, and to whether the simulators used to collect samples are stochastic or deterministic. This is especially convenient, since in robotics deterministic controllers are used widely, while the Reinforcement Learning community frequently considers stochastic policies and environments. With our model: the stochasticity of either environment or controllers (or both) will be encoded in $p(\btau | \bx)$. Even in the case of deterministic controllers and environment (deterministic relationship between $\bx$ and $\bxi$) the model remains meaningful because of the bottleneck $\btau$ and randomness coming from sampling of $\bx$ during data collection.

One fair question would be: why learn an embedding into lower-dimensional space of path $\btau$ jointly with learning $p(\btau|\bx)$, instead of decomposing the problem into separate dimensionality reduction and $p(\btau|\bx)$ modeling stages.
For an arbitrary space of paths  (either low-dimensional, or even the original high-dimensional space of trajectories) the relationship between $\bx$ and the corresponding probability distribution over the paths would be challenging. This is because it involves the controller properties and the dynamics of the physical environment, both of which are usually highly non-trivial. In the joint model we propose,  
$p(\btau|\bx)$ term can be seen as a `regularization' part of the ELBO. It keeps the latent representation s.t. it is well suited for modeling the relationship between $\bx$ and $\btau$. The terms pertaining to `reconstructing' original trajectories are the encoder $q(\btau|\bxi)$ and decoder  $p(\bxi|\btau)$. Learning progress for these is fast if there is sufficient capacity in the bottleneck $\btau$. However, if these make fast progress, but learn the space of latent paths $\btau$ that is not easy to relate to the space of controllers -- then $p(\btau|\bx)$ will drop. Hence ELBO will be lower for this `inconvenient' representation for $\btau$, encouraging alternatives. Consequently, our joint representation allows not only `compressing' the space of trajectories, but also finding a compression scheme that simplifies the problem of modeling $p(\btau|\bx)$.

As is customary with VAEs, at first we expressed variational approximate posterior and generative model components by multivariate Gaussians with diagonal covariance. Later we found that using Laplace distributions yielded more consistent training results. The reconstruction and generation for successful training runs were comparable. However, some runs using Gaussians collapsed to the mean instead of learning useful latent representations. So we kept Laplace as the default choice.

\section*{Appendix B: SVAE-DC Training Visualizations}
Below we include visualizations of the training progress. We developed an easy-to-use training pipeline that generates training statistics and visualization videos in Tensorboard. We included our code and a detailed README with instructions on how to install and use the codebase. We took care to comment our implementation, so all further details about our implementation, parameter choices and training procedure would be easy to infer from the code attached to this submission.

\begin{figure}[H]
    \centering
    \includegraphics[width=0.9\textwidth]{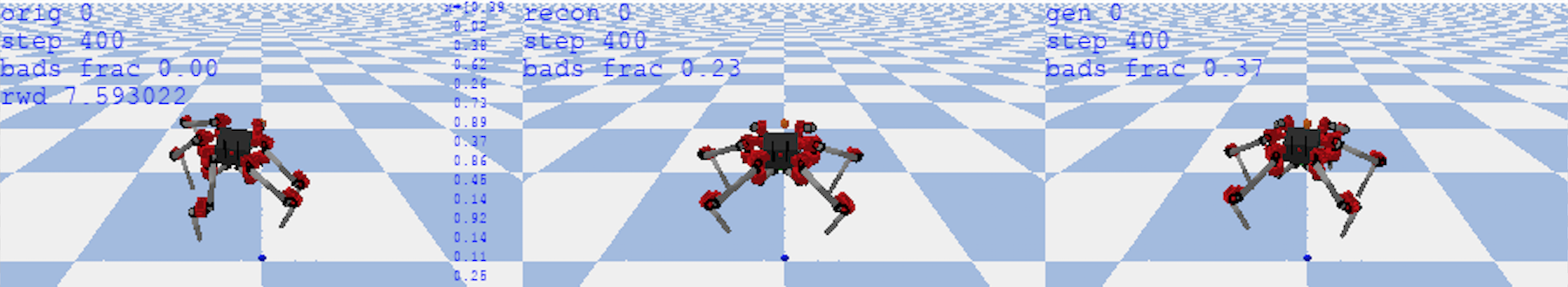}
    \vspace{4px}
    \includegraphics[width=0.9\textwidth]{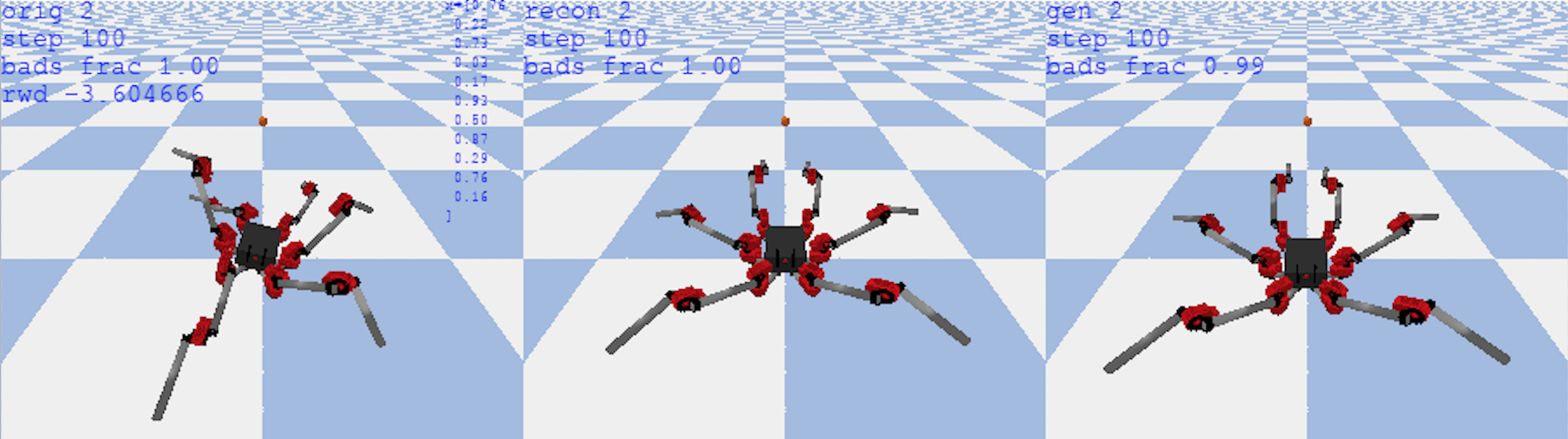}
    \caption{SVAE-DC training progress on Daisy. See full description in Figure~\ref{fig:svae_training_viz} caption on the next page.}
\end{figure}

\begin{figure}[H]
    \centering
    \vspace{-15px}
    \includegraphics[width=0.9\textwidth]{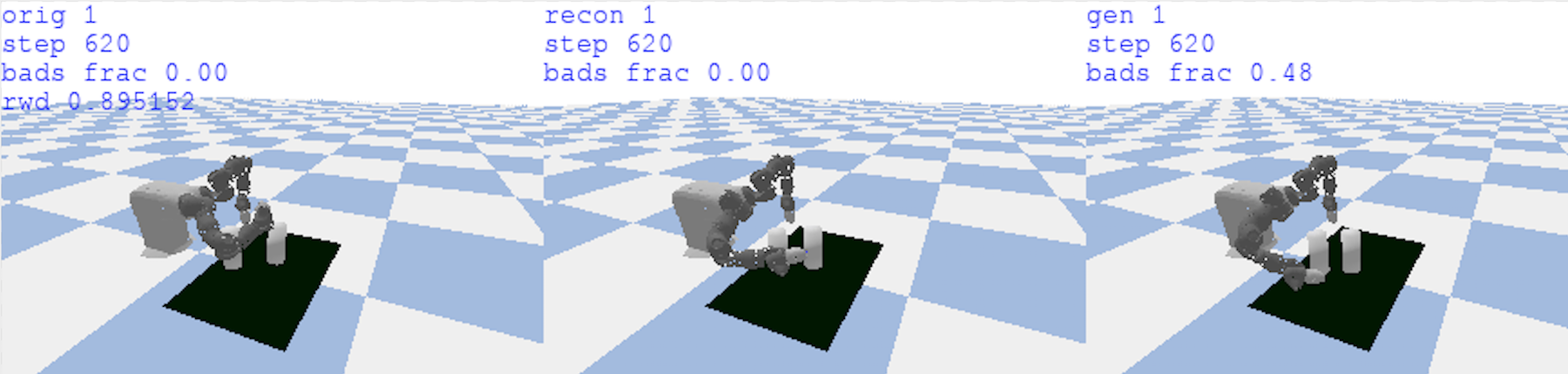}
    \vspace{4px}
    \includegraphics[width=0.9\textwidth]{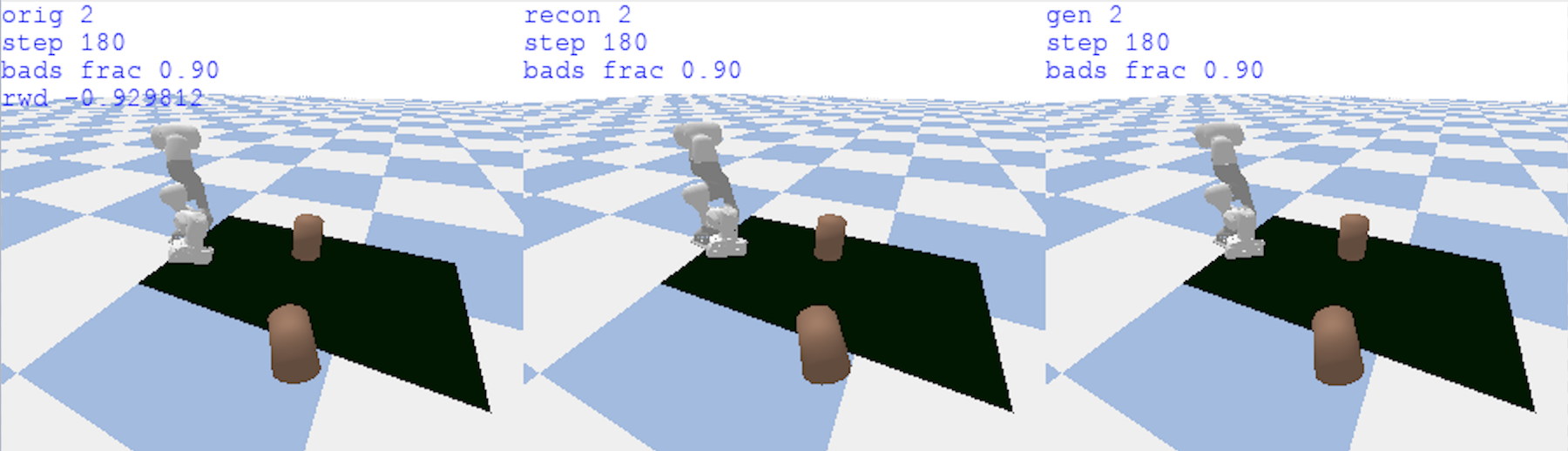}
    \caption{SVAE-DC training progress on Yumi (middle) and Franka Emika (bottom) environments. Observations consist of robot joint angles, object positions and orientations sampled at 500Hz in pybullet simulator. Figures are obtained by visualizing the original (left) reconstructed (middle) and generated (right) state of the robot and objects in the simulator (in this work we are not considering visual observations, so not learning to process pixel data). `Bads frac' indicates what percentage of the original trajectory is spent in undesirable parts of the space (i.e. $y$). During generation this is obtained by using $\psi$ part of the latent samples as input to $p(y|\psi)$ NN. For successful BO, perfect $y$ fit is not needed. For a useful separation, it is enough for SVAE-DC to fit $y$ for bad controllers and to report approximate lower $y$ for the good ones.}
    \label{fig:svae_training_viz}
    \vspace{-10px}
\end{figure}

\section*{Appendix C: Parametric vs Intrinsic Dimensionality}
\vspace{-7px}
When optimizing higher-dimensional controllers with few trials, one could question whether BO with SE kernel should be among the baselines. If our reward functions came from an arbitrary distribution, for BO in 30D space, for example, we would expect to need at least 60 trials to starts seeing the benefits. However, our reward landscapes come from real-world problems, not from purely analytic constructions. While robotics problems have a clear parametric dimensionality, their intrinsic dimensionality is usually unknown. The vision community is familiar with this concept: they frequently refer to a `lower-dimensional manifold of real-world images'. The intrinsic dimensionality of vision problems could be orders of magnitude lower than their parametric dimensionality expressed in pixel space.

\begin{wrapfigure}{r}{0.22\textwidth}
\vspace{-10px}
\includegraphics[width=0.22\textwidth]{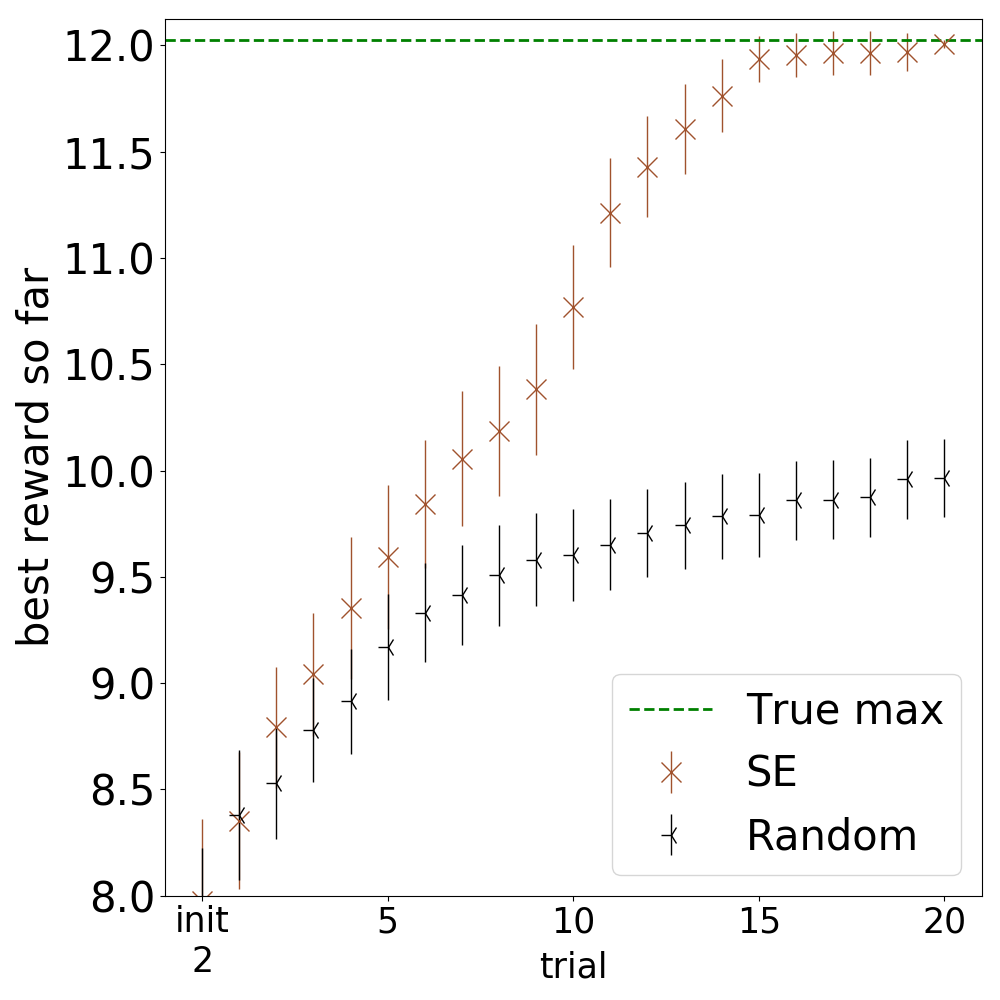}
\caption{BO in 30D when only 3 dimensions contribute significantly.}
\vspace{-15px}
\label{fig:bo_basic_intrd3}
\end{wrapfigure}
In the context of BO, consider a 30D quadratic: $f(\bx)\!=\!\sum_i (x_i\!+\!1)^2, \ \bx\!\!\in\!\!\R^{30}$ with $x_i \!\in\![0,1]$. Even on this simple quadratic BO with SE kernel gives only modest gains for the first 60 trials. Now consider $f$ such that a large number of dimensions do not contribute significantly: $f(\bx)\!=\!\sum_{i=1}^3 (x_i\!+\!1)^2 + 0.001\sum_{i=4}^{30} x_i$. Figure~\ref{fig:bo_basic_intrd3} shows that BO with SE kernel succeeds with few trials (as long as hyperparameters do not force BO to over-explore). So SE baseline is a reasonable check for adequate performance in such settings.

In most cases it is difficult to estimate (or even approximately guess) the intrinsic dimensionality of a real-world problem. Parametric representation doesn't even give an upper bound on complexity. In robotics,  intrinsic dimensionality of a problem could be higher than parametric dimensionality of its most commonly used representation. For example, the effects of friction are sometimes abstracted away as a few parameters of a simplified friction model. When such problems are declared `low-dimensional', it creates a misconception that they are `easy'. In fact, they remain hard for cases where friction matters for success and a crude model is inadequate.
It is also not easy to gauge the complexity of a problem by applying algorithms like PCA on the whole optimization space. A `simple' structure might be characteristic for only a part of the space. For example, failing  controllers could exhibit near chaotic behavior, making $f(\bx)$ have high intrinsic dimensionality when considering the whole space. But this space might contain sub-regions, where the relationship between reward and change in controller parameters is gradual. If domain knowledge or informed kernels can help find a few points close to a successful region: the gains for BO could be paramount. BO could quickly focus exploration on the promising regions, without being restricted to a particular model or simulator structure that originally helped to point to a promising part of the space.

\end{document}